\documentclass[11pt]{article}

\usepackage[T1]{fontenc}
\usepackage[utf8]{inputenc}
\usepackage{lmodern}
\usepackage{microtype}
\usepackage[margin=1in]{geometry}

\usepackage{csquotes}
\usepackage{graphicx}
\usepackage{booktabs}
\usepackage{threeparttable}
\usepackage{setspace}
\usepackage{authblk}
\usepackage{hyperref}
\usepackage{hyphenat}

\usepackage[
  backend=biber,
  style=apa,
  natbib=true,
  sorting=nyt
]{biblatex}
\DeclareLanguageMapping{american}{american-apa}
\title{Do LLM-Driven Agents Exhibit Engagement Mechanisms?\\
Controlled Tests of Information Load, Descriptive Norms, and Popularity Cues}

\author[1]{Tai-Quan Peng}
\author[2]{Yuan Tian}
\author[2]{Songsong Liang}
\author[3]{Dazhen Deng}
\author[2]{Yingcai Wu}

\affil[1]{Department of Communication, Michigan State University, USA}
\affil[2]{College of Computer Science and Technology, Zhejiang University, China}
\affil[3]{School of Software Technology, Zhejiang University, China}

\date{}

\begin{document}

\maketitle

\begin{center}
\small
Correspondence: Dr.\ Tai-Quan Peng, Department of Communication, Michigan State University, East Lansing, MI 48824, USA.\\
Email: \href{mailto:winsonpeng@gmail.com}{winsonpeng@gmail.com}.
\end{center}

\begin{abstract}
Large language models make agent-based simulation more behaviorally expressive, but they also sharpen a basic methodological tension: fluent, human-like output is not, by itself, evidence for theory. We evaluate what an LLM-driven simulation can credibly support using information engagement on social media as a test case. In a Weibo-like environment, we manipulate information load and descriptive norms, while allowing popularity cues (cumulative likes and Sina Weibo--style cumulative reshares) to evolve endogenously. We then ask whether simulated behavior changes in theoretically interpretable ways under these controlled variations, rather than merely producing plausible-looking traces. Engagement responds systematically to information load and descriptive norms, and sensitivity to popularity cues varies across contexts, indicating conditionality rather than rigid prompt compliance. We discuss methodological implications for simulation-based communication research, including multi-condition stress tests, explicit no-norm baselines because default prompts are not blank controls, and design choices that preserve endogenous feedback loops when studying bandwagon dynamics.
\end{abstract}


\section{Introduction}

Computational simulation has been increasingly used to study communication processes that are difficult to observe directly \citep{waldherrWorldsAgentsProspects2021}. Large language models (LLMs) now enable researchers to construct agents that can consume information \citep{tornbergSimulatingSocialMedia2023}, interpret messages and context \citep{loruSimulationJudgmentLLMs2025}, and generate human-like responses and behavioral decisions at scale \citep{xieUsingLargeLanguage2025}, opening new possibilities for controlled experimentation on human behaviors and social interactions \citep{bail_can_2024}. However, methodological discussions caution that behavioral resemblance alone is insufficient grounds for inference. LLM-driven agents cannot be assumed to instantiate the same underlying mechanisms as human subjects, and their outputs require validation against the theoretical processes they are meant to model \citep{kozlowskiSimulatingSubjectsPromise2025}. As AI-generated behavior becomes increasingly fluent and contextually appropriate, the central question shifts from plausibility to mechanism: do LLM-driven agents reproduce theoretically grounded communication processes, or do they merely generate outputs that resemble them?

Recent work underscores this concern. Although LLMs can produce human-like output, they may not engage in human-like reasoning \citep{demszky_using_2023}. Large-scale replication efforts further suggest that LLM-based simulations can inflate effect sizes or generate patterns inconsistent with original findings \citep{cui_large-scale_2025}. These findings indicate that the validity of LLM-driven simulation cannot be assumed but must be demonstrated. Rather than evaluating simulations in terms of surface realism, the methodological challenge is to test whether established cognitive or social mechanisms emerge under controlled manipulation in LLM-driven simulations \citep{li_simulating_2025}. 

This study addresses this challenge by using social media information engagement as a theoretically grounded test case. Engagement decisions on social media are widely understood to reflect informational constraints, behavioral norms, and visible social feedback \citep{david_easley_networks_2010}, yet these mechanisms are difficult to isolate because exposure volume, foregone alternatives, and algorithmic visibility are typically unobserved and endogenous. We therefore employ an LLM-based social media simulation that experimentally varies information load and descriptive normative context while allowing social feedback signals to evolve endogenously. By separating participation thresholds from engagement allocation, we examine whether theory-implied regularities emerge and interact in ways consistent with widely tested theories of attention, social influence, and normative behavior. In this sense, the simulation is designed to serve theoretical purposes: we treat LLM agents as methodological instruments for testing engagement mechanisms under controlled conditions, while also evaluating what kinds of theoretical inference such simulation can credibly support.

\section{Theoretical Framework and Research Questions}

Research on social media engagement emphasizes three interrelated forces: informational load, behavioral norms, and popularity cues. Users allocate attention under conditions of limited cognitive resources, form expectations about what behaviors are typical, and observe visible signals indicating which content others have already responded to. Engagement therefore reflects not only message content or network structure, but also how individuals navigate the informational and social context in which choices are made.

Although each of these mechanisms is well established, isolating their independent and joint effects remains difficult in observational research. On digital platforms, information load, normative signals, and popularity cues are often endogenous to algorithmic recommendation systems and user networks. Digital trace data record realized actions, but they rarely capture the full set of posts to which users were exposed, the alternatives they ignored, or the ichoice-set conditions under which decisions were made \citep{salganikBitBitSocial2019}. The present study uses controlled simulation to examine whether theory-implied regularities associated with information load, descriptive norms, and bandwagon effects emerge when these contextual features are systematically varied, while allowing popularity cues to evolve endogenously.

\paragraph{Engagement as a Two-Stage Outcome}

Engagement is not a single, undifferentiated act. When encountering content on social media, individuals decide whether to respond at all and, conditional on responding, which engagement form to select. We therefore treat engagement as a two-stage outcome. The first stage concerns the participation threshold---whether an individual engages with a post or remains a passive reader. The second stage concerns engagement allocation---the selection among available response forms that vary in effort and visibility.

This distinction clarifies how different mechanisms may operate. Information load is expected to shape participation thresholds by increasing competition among alternatives. Descriptive norms are expected to shift engagement allocation by altering which response forms are treated as typical. Popularity cues may influence both stages by signaling prior validation, thereby generating bandwagon-consistent amplification.

\paragraph{Information Load and Attention Constraints}

Information load has long been understood as a fundamental constraint on information processing. Classic psychological research demonstrates that individuals possess limited cognitive capacity, such that increases in input volume alter attention allocation and depth of processing \citep{miller_magical_1956}. This capacity-based perspective has been extended to mediated communication contexts through models such as the Limited Capacity Model of Motivated Mediated Message Processing, which conceptualizes media users as resource-limited processors who must allocate finite resources across concurrent informational demands \citep{lang_limited_2000}. In social media environments characterized by abundant exposure, higher information load is therefore expected to increase competition among posts, raise engagement thresholds, and encourage more selective responding \citep{rodriguez_quantifying_2014, weng_competition_2012, feng_competing_2015}.

In this study, information load is operationalized as the number of posts presented to agents at each decision point. This manipulation varies the size of the choice set at the moment of decision, allowing us to assess whether load-sensitive engagement regularities---especially reduced participation under higher load---emerge within an LLM-driven simulation. The focus is not on attributing human cognitive limitations to artificial agents, but on testing whether established load-related behavioral regularities are reproduced under controlled variation in choice-set size.

\paragraph{Bandwagon Effects and Popularity Cues}

Bandwagon effects refer to individuals' tendency to rely on visible indicators of others' behavior when making decisions. On social media platforms, likes, reposts, and quotes provide publicly observable \textit{popularity cues} that signal attention and endorsement. Classic models of informational influence and cascade dynamics suggest that individuals may infer content value from others' actions, even when private evaluations are available \citep{bikhchandani_theory_1992, david_easley_networks_2010}. Empirical research further shows that early popularity cues can causally shape subsequent engagement, generating self-reinforcing dynamics consistent with cumulative advantage \citep{muchnik_social_2013}. These findings imply that popularity cues lower participation thresholds and increase the likelihood of subsequent response.

In the present simulation, agents observe dynamically updated engagement counts before making decisions. Unlike information load and descriptive norms, popularity cues are not experimentally imposed but emerge endogenously from prior interactions. This design allows bandwagon processes to unfold within the simulated environment and enables tests of whether responsiveness to popularity cues depends on information load and descriptive norm regimes.

\paragraph{Descriptive Norms and Engagement Allocation}
Engagement behavior is shaped not only by informational constraints but also by perceptions of what others typically do. Social psychological research distinguishes between descriptive norms, which reflect common behavioral patterns, and injunctive norms, which reflect perceived approval or disapproval \citep{cialdini_focus_1990}. The present study focuses on descriptive norms. By providing prevalence information about which engagement forms are typical in a given environment, descriptive norms are expected to shift the relative likelihood of available response options.

In LLM-driven simulations, descriptive norms can be instantiated through prompting. Prompts that describe typical engagement patterns can be treated as manipulations of descriptive norms because they provide information about what others commonly do, without mandating a specific action. This approach parallels experimental designs in which researchers provide descriptive norm information to shift behavior by altering perceived prevalence and salience \citep{allcottSocialNormsEnergy2011, papakonstantinouSystematicReviewMetaanalysis2025}. Accordingly, the theoretical expectation is not that agents mechanically reproduce stated proportions, but that descriptive norm cues function as contextual information that biases engagement allocation within the choice process.

\subsection{Research Questions}

The present study examines engagement as a two-stage outcome: whether an agent engages rather than reads, and, conditional on engagement, which engagement form is selected. Information load, descriptive norms, and popularity cues are expected to shape these stages in distinct but interrelated ways. Rather than treating these mechanisms as purely substantive predictors, we evaluate whether LLM-driven agents reproduce theory-implied engagement regularities when these contextual features are systematically varied and when bandwagon-relevant popularity cues develop endogenously.

Accordingly, the study addresses the following questions:

\textbf{RQ1 (Engagement Threshold Validation):}  
To what extent do LLM-driven agents exhibit load-sensitive changes in the probability of engagement, consistent with capacity-based expectations?

\textbf{RQ2 (Bandwagon Sensitivity):}  
Do popularity cues increase the likelihood of engagement and amplify response probabilities in ways consistent with bandwagon effects?

\textbf{RQ3 (Norm-Consistent Allocation):}  
Do descriptive normative prompts systematically reorganize engagement allocation in directions consistent with the signaled behavioral expectations?

\textbf{RQ4 (Mechanism Interaction):}  
How do information load, descriptive norms, and popularity cues jointly shape engagement thresholds and engagement allocation within the simulated environment?

\section{Research Design}

This study employs an LLM-driven agent-based simulation to examine how \textit{information load}, \textit{descriptive norms}, and \textit{popularity cues} jointly shape engagement decisions in a Weibo-like environment. Sina Weibo shares core structural features with other microblogging systems, including X (formerly Twitter) and BlueSky: public follower-following networks, feed-based recommendation systems, short-form posts, visible engagement metrics, and interaction affordances such as like, repost (retweet), and quote-with-comment. The theoretical mechanisms under investigation are therefore not platform-specific. Using Sina Weibo provides empirical grounding while preserving structural comparability with other microblogging environments.

\subsection{Simulation Framework}

The simulation framework builds on the OASIS architecture \citep{yang_oasis_2024}, which integrates network structure, recommendation systems, and LLM-powered agents to emulate platform-like interaction processes. The framework consists of three interconnected modules. The \textit{environment module} initializes agent profiles, follower–following relationships, and seed posts. The \textit{recommendation module} selects the posts shown to each agent at each activation, combining algorithmic ranking based on post recency and topic relevance with content from followed accounts. The \textit{agent interaction module} uses an LLM to generate behavioral decisions based on posts shown, their associated popularity cues, the agent profile, and recent interaction history. \autoref{tab:design_components} summarizes the empirical inputs, experimental manipulations, controlled parameters, and simulated components.

\begin{table}[htbp]
\centering
\caption{Summary of Empirical Inputs, Experimental Manipulations, and Simulated Components}
\label{tab:design_components}
\begin{threeparttable}
\footnotesize
\setlength{\tabcolsep}{4pt}
\begin{tabular}{p{4.2cm} p{3.2cm} p{5.0cm}}
\toprule
\textbf{Study Component} & \textbf{Origin} & \textbf{Role in Design} \\
\midrule
Agent profiles ($n = 558$) & Real-world data (Sina Weibo) & Provide topic-specific user attributes and network structure \\
Follower--following relationships & Real-world data (partial network) & Establish diffusion pathways and content visibility channels \\
Seed posts ($n = 50$, Chinese) & Real-world data (DeepSeek discussions) & Initialize cascades with empirically grounded content \\
\midrule
Information load (4 conditions) & Experimentally manipulated & Vary choice-set size at each activation (posts shown) \\
Descriptive norm regime (3 conditions) & Experimentally manipulated & Manipulate perceived engagement prevalence via prompt framing \\
\midrule
Agent activation probability (1\% per timestep) & Controlled simulation parameter & Hold activity rate constant across conditions \\
Network size (558 agents) & Controlled simulation parameter & Hold participation pool constant across conditions \\
Recommendation logic & Controlled simulation parameter & Maintain consistent content selection and ranking rules \\
\midrule
Popularity cues (cumulative engagement counts) & Endogenous (simulated) & Generate dynamic popularity cues that can produce bandwagon effects \\
Agent decisions (read, like, repost, quote) & LLM-generated & Produce engagement behavior under controlled conditions \\
Repost and quote posts & LLM-generated & Extend cascades through redistribution and commentary \\
\bottomrule
\end{tabular}

\begin{tablenotes}
\vspace{0.3em}
\footnotesize
\item Note. Popularity cues were not experimentally manipulated; cumulative engagement counts evolved endogenously within each simulation run.
\end{tablenotes}
\end{threeparttable}
\end{table}

\subsection{Inputs: Seed Posts, Agents, and Network Initialization}

To ensure topical coherence, we collected 50 unique seed posts centered on DeepSeek-related discussions from Sina Weibo. All posts were written in Chinese and ranged between 150 and 300 tokens, with duplicate content removed. At the start of each simulation run, these posts were injected into the environment and attributed to eight influential agents selected based on real-world indicators (e.g., number of follower, verified accounts). These seed posts served as the initial sources for subsequent engagement.

The simulation included 558 agents whose profiles and follower–following relationships were derived from real users engaged in DeepSeek-related discussions. Active users were defined as those who had performed at least one primary interaction (posting, reposting, or quoting) within the topic domain. This sampling strategy preserves topic-specific participation patterns while holding network structure constant across experimental conditions.

Each agent was powered by the Qwen-8B language model \citep{yangQwen3TechnicalReport2025} (temperature = 0.6). The model choice aligns with the linguistic and cultural context of the platform. The interaction prompt included (a) the posts shown to the agent together with popularity cues, (b) the agent’s profile information, and (c) the agent’s eight most recent interactions.

System prompts and agent instructions were written in English, whereas seed posts and simulated platform content were in Chinese. We used English prompts to keep the wording of instructions consistent across conditions and to avoid introducing unintended variation in the descriptive norm manipulation, while retaining Chinese content to preserve the Weibo-like linguistic context. Since both the prompt language and content language are held constant across all conditions, the comparisons across information load, descriptive norms, and popularity cues are driven by the experimental manipulations rather than by differences in language framing.

\subsection{Simulated Components: Engagement Behaviors}

At each activation, agents selected one of four behaviors: read, like, repost, or quote. Reading was treated as passive exposure and not coded as active engagement. Liking represents low-effort endorsement, increasing visible popularity metrics without modifying the original content. Reposting and quoting both generate new posts that re-enter the recommendation system, but they differ structurally. A repost creates a new post that repeats the original content while attributing it to the reposter’s account, mirroring Sina Weibo’s repost function. A quote produces a new post that includes agent-generated commentary in addition to the original content. This distinction differentiates redistribution without modification from redistribution accompanied by added expression. All actions were logged with timestamps and source-link identifiers.

Two clarifications are warranted. First, although engagement is conceptualized analytically as a two-stage outcome—distinguishing participation thresholds from engagement allocation—agents were not required to make sequential decisions in the simulation. At each activation, the model selected a single action from the full set. The two-stage structure is introduced at the level of analysis to separate engage-versus-read decisions from the distribution of engagement forms, not to impose a cognitive sequence.

Second, commenting was not included as a focal behavior. While comments contribute to conversational interaction, they typically remain embedded within comment threads and do not generate new feed items. In contrast, likes directly update popularity cues, and reposts and quotes generate new posts that can circulate in additional feeds. Because the study examines mechanisms shaping engagement likelihood and engagement allocation in visibility-mediated systems, the analysis focuses on behaviors that directly affect popularity cues or content propagation.

\subsection{Experimental Manipulations and Endogenous Features}

The study employs a factorial design in which information load and descriptive norm regime were experimentally manipulated, while network structure, activation probability, model parameters, and recommendation logic were held constant. Popularity cues were present in all conditions but were not experimentally varied.

\paragraph{Information Load (Manipulated).}
Information load was operationalized by varying the size of the post set shown to an agent at each activation, while preserving a constant baseline of content from followed accounts. In each activation, agents always received three posts drawn from accounts they followed, reflecting routine feed exposure through network ties. Information load was manipulated by varying the number of additional posts supplied by the recommendation algorithm: 4 algorithmic posts in the lowest-load condition (total $=7$ posts), 7 algorithmic posts in the low-load condition (total $=10$), 15 algorithmic posts in the medium-load condition (total $=18$), and 30 algorithmic posts in the high-load condition (total $=33$). The posts shown could include seed posts as well as repost and quote posts generated during the simulation.

This design isolates the effect of information load as choice-set expansion driven by algorithmic recommendation, while holding constant the minimal level of network-based exposure. Keeping followed-account content fixed ensures that differences across conditions reflect changes in the volume of algorithmically supplied alternatives rather than changes in whether agents receive content from their social ties.

\paragraph{Behavioral Norm Regimes (Manipulated).}
We manipulated descriptive norms via the system prompt. In the \textit{no-norm condition}, agents received no explicit prevalence information about engagement forms. In the \textit{like-dominant condition}, the prompt described liking as the typical default (approximately like 80\%, repost 15\%, quote 5\%). In the \textit{repost-dominant} condition, the prompt described reposting as the typical default among redistribution behaviors (approximately repost 90\%, quote 10\%), without assigning an explicit probability to liking, reflecting the conceptual emphasis on redistribution rather than endorsement.

The numeric values embedded in the prompts served as heuristic prevalence cues rather than strict behavioral targets. No algorithmic constraints enforced proportional compliance, and all engagement options remained available in every condition. Importantly, the prompts described the relative prevalence of engagement forms (like, repost, quote) but did not prescribe overall participation rates relative to reading. This design targets engagement allocation without directly specifying engagement frequency.

\paragraph{Popularity Cues (Endogenous Feature).}
In all conditions, agents observed popularity cues for each post prior to making a decision, including cumulative likes and cumulative reshares. Cumulative reshares follow Sina Weibo’s native convention of reporting redistribution as a combined count that aggregates reposts and quotes. These counts were dynamically updated and evolved endogenously through prior agent interactions. Because popularity cues were not experimentally imposed, the simulation allows bandwagon effects to unfold as an emergent consequence of accumulated engagement.

\subsection{Simulation Procedure}

Each simulation represented a 24-hour period divided into 480 timesteps (3-minute intervals). At each timestep, agents were activated with a probability of 1\%, yielding approximately five active agents per interval. Agents were not active simultaneously, reflecting the asynchronous nature of social media participation. Activation probability was held constant across all conditions. Fixing the proportion of active users per interval controls variation in participation intensity and ensures that differences in engagement likelihood and engagement allocation are attributable to experimental manipulations rather than changes in effective user activity.

Activated agents received a set of posts determined by their assigned information-load condition and generated a behavioral response based on post content, popularity cues, and descriptive norm regime. Agent actions updated the environment in real time by generating reposts and quotes and updating cumulative engagement counts. Through repeated iterations, engagement patterns and visibility distributions emerged endogenously from interaction.

\subsection{Analytical Design}
To evaluate engagement as a two-stage outcome, we estimated two complementary regression models corresponding to participation thresholds and engagement allocation.

In the first stage, we modeled whether an agent engaged with a post (like, repost, or quote) or remained a passive reader using binary logistic regression. Key predictors include information load, descriptive norm regime, and popularity cues. For parsimony, we operationalized popularity cues as a single composite score based on a post's prior cumulative engagement, combining cumulative likes and cumulative reshares and applying a log transformation to reduce skew. This aggregation treats popularity as an overall visibility signal rather than separating endorsement from redistribution and avoids multicollinearity between highly correlated engagement counts. Although the descriptive norm manipulation was designed to shift engagement allocation rather than participation thresholds, it is included to assess whether prevalence information indirectly alters overall engagement likelihood and whether information load and bandwagon effects vary across norm regimes. This model primarily addresses RQ1 and contributes to RQ2 and RQ4.

In the second stage, conditional on engagement, we examined allocation among liking, reposting, and quoting using multinomial logistic regression. Predictors again include information load, descriptive norm regime, and the same composite popularity cue, along with their interactions. Here the descriptive norm manipulation is central: the analysis evaluates whether prevalence framing reorganizes the relative probability of engagement forms and whether these shifts depend on information load and popularity cues. This stage directly addresses RQ3 and also informs RQ2 and RQ4 by testing how bandwagon effects and information load condition engagement allocation.

Together, these models separate engage-versus-read decisions from the distribution of engagement forms. This analytical structure enables tests of whether information load, descriptive norms, and popularity cues operate uniformly across stages or exhibit stage-specific effects, and whether their joint influence reveals interaction patterns consistent with bandwagon dynamics and context-dependent engagement.

\section{Results}
\subsection{Descriptive Findings}

We first verify the realized \textit{information load} across experimental conditions. Although the number of posts shown at each activation was specified by design, minor variation occurred because the recommendation process is dynamic. Across descriptive norm regimes, the mean number of algorithmically selected posts closely matched the intended levels. In the lowest-load condition, the average number of algorithmic posts ranged from 4.45 to 4.64; in the low-load condition, from 7.47 to 7.67; in the medium-load condition, from 15.4 to 15.7; and in the high-load condition, from 30.4 to 30.8. Within each condition, the observed minimal and maximal indicate modest fluctuation (e.g., the high-load condition ranged from 30 to 35 algorithmic posts). Overall, the realized distributions clustered around the intended thresholds, indicating that information load was successfully differentiated across the four conditions. Accordingly, information load is treated as a four-level categorical variable (lowest, low, medium, high) in subsequent analyses.

We next describe the distribution of response types (read, like, repost, quote) across the 12 combinations of information load and descriptive norm regime. Across all conditions, reading accounted for the majority of actions, ranging from 73.1\% (like-dominant descriptive norm under lowest information load) to 98.0\% (repost-dominant descriptive norm under high information load), indicating that active engagement constituted a minority of decisions throughout the simulation.

\begin{figure}
    \centering
    \caption{Observed distribution of engagement behaviors across normative regimes and information load levels.}
    \includegraphics[width=1\linewidth]{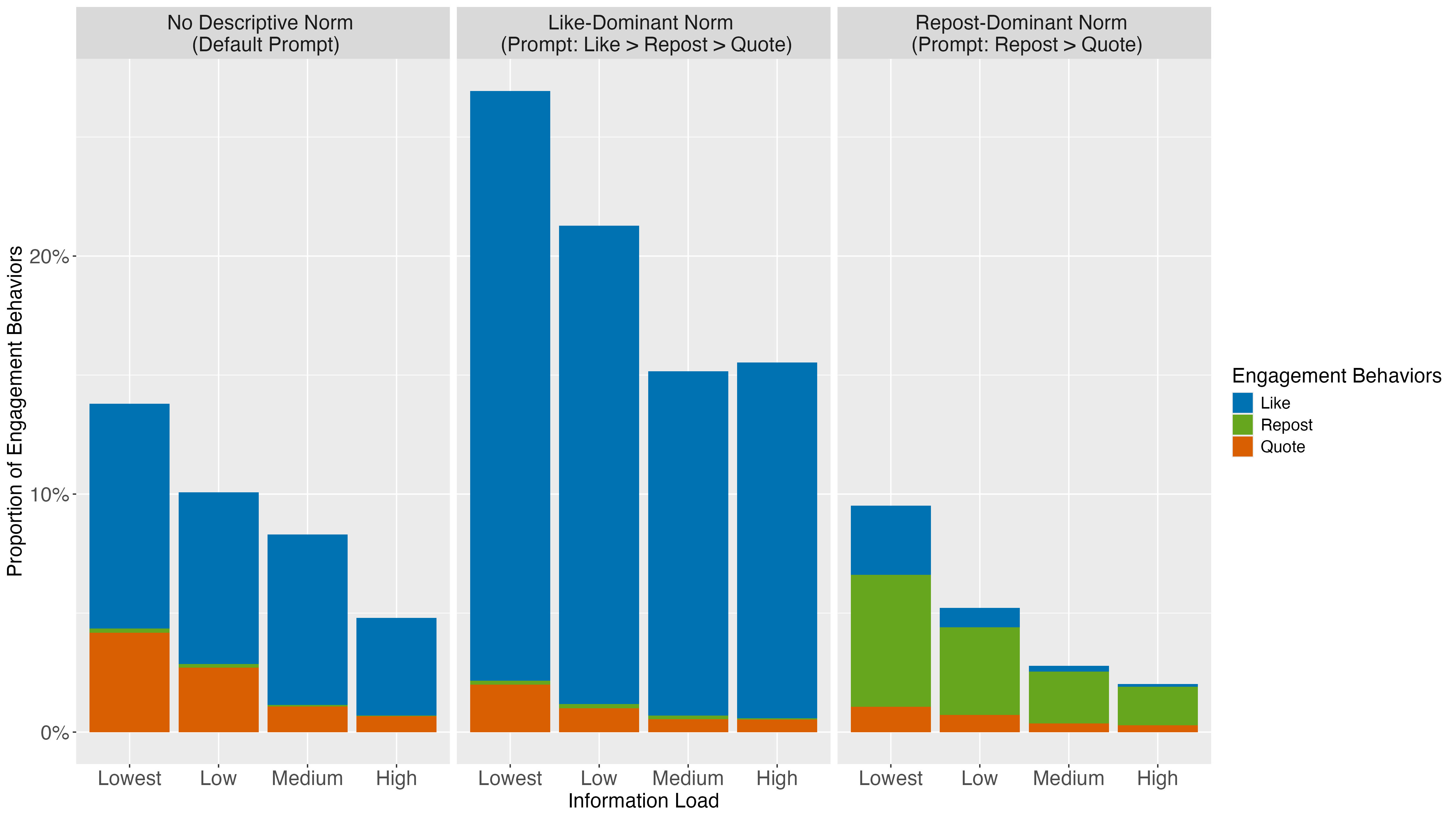}
    \label{fig:engagement_distribution}
\end{figure}

\autoref{fig:engagement_distribution} summarizes the relative shares of the three engagement forms across conditions. Under the no-norm condition, liking ranged from 4.1\% to 9.5\% across information-load levels, quoting ranged from 0.7\% to 4.2\%, and reposting remained rare (below 0.2\%). Under the like-dominant descriptive norm, liking was consistently the most prevalent engagement form (14.5\% to 24.8\%), whereas quoting and reposting remained comparatively infrequent (quoting 0.5\% to 2.0\%; reposting below 0.2\%). Under the repost-dominant descriptive norm, reposting became substantially more prevalent (1.6\% to 5.5\%), while liking remained minimal (0.1\% to 2.9\%) and quoting remained low (0.3\% to 1.1\%), particularly under medium and high information load.

\subsection{Engagement Threshold (Engage vs. Read)}

We examined whether agents chose to engage with a post (like, repost, or quote) rather than read only using binary logistic regression. The full model significantly improved fit relative to the intercept-only model, $\chi^2$(23) = 141{,}852.53, $p < .001$, and demonstrated substantial explanatory power (McFadden $R^2$ = 0.496; AIC = 144{,}132). Estimates are reported in \autoref{tab:logistic}.

\begin{table}[htbp]
\centering
\begin{threeparttable}
\caption{Binary logistic regression of engagement on prior cumulative engagement, information load, and descriptive norm regime}
\label{tab:logistic}
\footnotesize
\setlength{\tabcolsep}{4pt}
\begin{tabular}{p{7.2cm}rrrr}
\toprule
Predictor & $B$ & $SE$ & Odds Ratio & $p$-value \\
\midrule
Intercept & -2.787 & 0.041 & 0.062 & $< .001$ \\
Popularity Cues \textsuperscript{a} & 4.624 & 0.115 & 101.871 & $< .001$ \\

Low Load (vs. Lowest Load)\textsuperscript{b} & -0.438 & 0.056 & 0.645 & $< .001$ \\
Medium Load (vs. Lowest Load)\textsuperscript{b} & -0.758 & 0.051 & 0.469 & $< .001$ \\
High Load (vs. Lowest Load)\textsuperscript{b} & -1.297 & 0.049 & 0.273 & $< .001$ \\

Like-dominant Norm (vs. No Norm)\textsuperscript{c} & 0.291 & 0.057 & 1.338 & $< .001$ \\
Repost-dominant Norm (vs. No Norm)\textsuperscript{c} & -0.465 & 0.064 & 0.628 & $< .001$ \\

\midrule
\multicolumn{5}{l}{\textit{Two-way interactions}} \\
\midrule
Popularity $\times$ Low Load & -0.066 & 0.146 & 0.936 & .653 \\
Popularity $\times$ Medium Load & -1.156 & 0.123 & 0.315 & $< .001$ \\
Popularity $\times$ High Load & -0.808 & 0.122 & 0.446 & $< .001$ \\

Popularity $\times$ Like-dominant Norm & -0.159 & 0.143 & 0.853 & .268 \\
Popularity $\times$ Repost-dominant Norm & 0.525 & 0.178 & 1.691 & .003 \\

Low Load $\times$ Like-dominant Norm & -0.011 & 0.077 & 0.989 & .887 \\
Medium Load $\times$ Like-dominant Norm & -0.013 & 0.070 & 0.987 & .850 \\
High Load $\times$ Like-dominant Norm & 0.472 & 0.066 & 1.603 & $< .001$ \\

Low Load $\times$ Repost-dominant Norm & -0.182 & 0.090 & 0.834 & .044 \\
Medium Load $\times$ Repost-dominant Norm & -0.425 & 0.084 & 0.654 & $< .001$ \\
High Load $\times$ Repost-dominant Norm & -0.298 & 0.080 & 0.742 & $< .001$ \\

\midrule
\multicolumn{5}{l}{\textit{Three-way interactions}} \\
\midrule
Popularity $\times$ Low Load $\times$ Like-dominant Norm & -0.279 & 0.180 & 0.757 & .121 \\
Popularity $\times$ Medium Load $\times$ Like-dominant Norm & -0.126 & 0.154 & 0.882 & .412 \\
Popularity $\times$ High Load $\times$ Like-dominant Norm & -0.711 & 0.151 & 0.491 & $< .001$ \\

Popularity $\times$ Low Load $\times$ Repost-dominant Norm & 2.639 & 0.286 & 13.995 & $< .001$ \\
Popularity $\times$ Medium Load $\times$ Repost-dominant Norm & 3.326 & 0.243 & 27.815 & $< .001$ \\
Popularity $\times$ High Load $\times$ Repost-dominant Norm & 3.649 & 0.226 & 38.427 & $< .001$ \\
\bottomrule
\end{tabular}

\begin{tablenotes}
\vspace{0.6em}
\footnotesize
\item Note. Dependent variable coded 1 = engagement (like, repost, or quote) and 0 = read only. $p$ values are two-tailed.
\item[\textsuperscript{a}] Popularity cues is operationalized as the log-transformed cumulative engagement a post had accumulated prior to the focal decision.
\item[\textsuperscript{b}] Information load reflects the number of posts shown to agents; the reference category is the lowest-load condition.
\item[\textsuperscript{c}] Descriptive norm regimes specify prevalence information about engagement forms; the reference category is the no-norm condition.
\end{tablenotes}

\end{threeparttable}
\end{table}

Popularity cues were strongly associated with higher engagement likelihood (B = 4.624, SE = 0.115, $p < .001$; OR = 101.87), consistent with bandwagon effects. Information load also showed a negative gradient in the baseline specification: relative to the lowest-load condition, the odds of engagement decreased under low load (OR = 0.65), medium load (OR = 0.47), and high load (OR = 0.27), all $p < .001$. Taken together, these main effects indicate that engagement is less likely when agents face larger choice sets, but more likely when posts carry stronger popularity cues.

\begin{figure}[htbp]
    \centering
    \includegraphics[width=\linewidth]{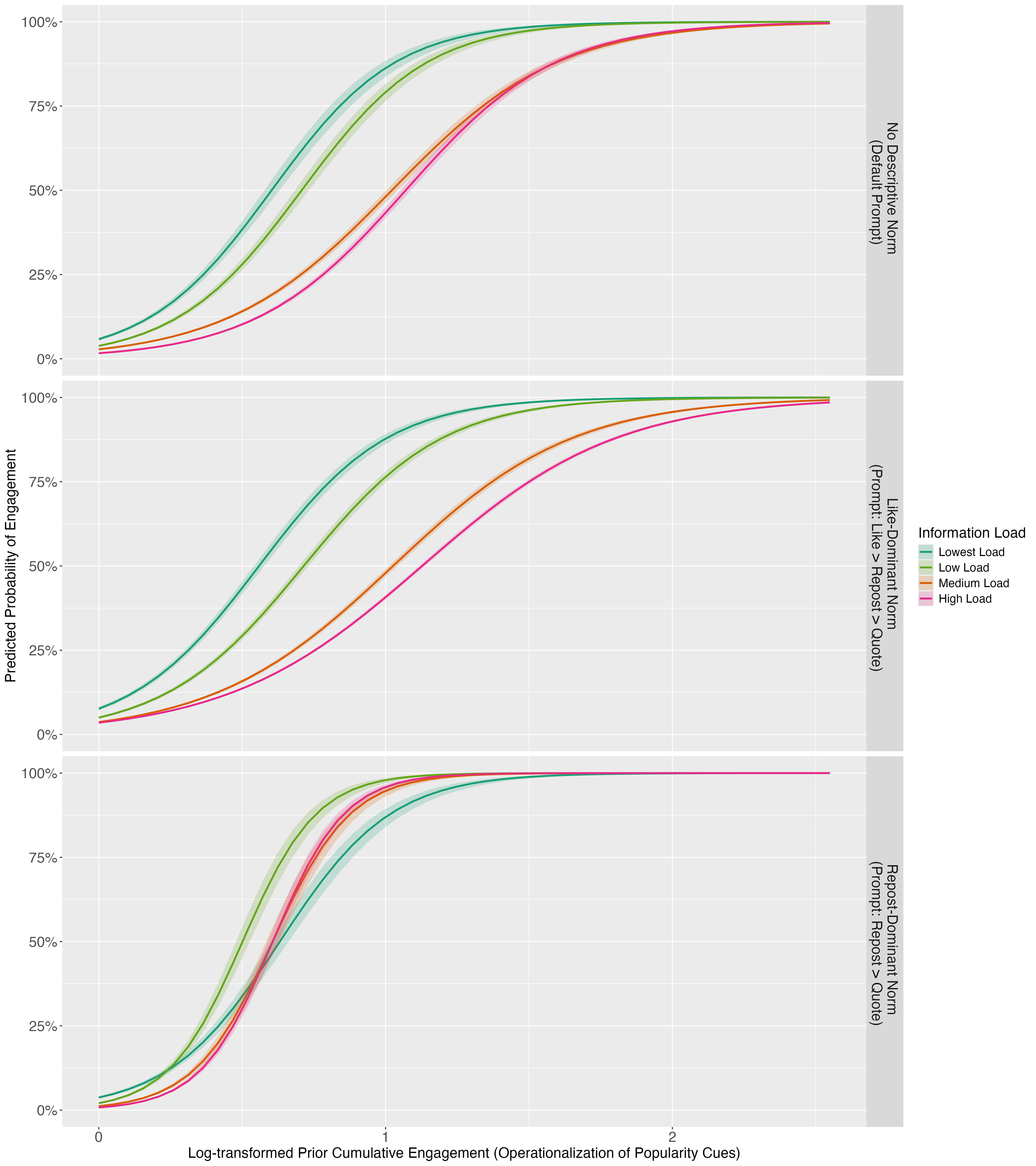}
    \caption{Predicted probability of engagement (vs.\ read-only) adjusted by information load, popularity cues, and normative regimes.}
    \label{fig:engagement_prob}
    
    \vspace{0.5em}
    \small
    \textit{Note.} Estimates are based on a logistic regression model including all two- and three-way interactions. Curves illustrate how the effects of popularity cues vary jointly with information load and normative regimes.
\end{figure}

The model also revealed a significant three-way interaction among popularity cues, information load, and descriptive norm regime (see \autoref{fig:engagement_prob}). It is worth noting that descriptive norm regimes were introduced to shift engagement allocation rather than participation thresholds. We therefore treat norm-related differences in the threshold model as informative boundary conditions rather than as the primary focus of the manipulation. With that caveat, two patterns are clear.

First, the no-norm and like-dominant norm regimes exhibit broadly similar load-sensitive structure. Holding popularity cues constant, the lowest information-load condition tends to show the highest engagement likelihood, with engagement probabilities declining as information load increases. The like-dominant regime shifts baseline engagement upward relative to no norm (B = 0.291, $p < .001$; OR = 1.34), but it does not fundamentally alter this monotonic ordering across information-load conditions over most of the popularity range.

Second, the repost-dominant regime exhibits a different shape. Baseline engagement likelihood is lower in this regime (B = -0.465, $p < .001$; OR = 0.63). However, responsiveness to popularity cues becomes substantially stronger as information load increases in the repost-dominant regime, as indicated by the positive interaction between popularity cues and repost-dominant norms (popularity cues $\times$ repost-dominant: B = 0.525, $p = .003$) and the large positive three-way interactions between popularity cues, information load, and repost-dominant norms (all $p < .001$; e.g., popularity cues $\times$ medium load $\times$ repost-dominant: B = 3.326; popularity cues $\times$ high load $\times$ repost-dominant: B = 3.649). Substantively, this implies a crossover pattern as illustrated in \autoref{fig:engagement_prob}: when popularity cues are weak, engagement is highest under the lowest information load and declines monotonically as information load increases; at moderate popularity-cue levels, the low-load condition can surpass the lowest-load condition; and at higher popularity-cue levels, the medium- and high-load conditions can surpass the lowest-load condition, with the lowest-load group becoming the least likely to engage.

Overall, the threshold results provide strong evidence for bandwagon effects and for load sensitivity in the baseline and like-dominant regimes (RQ1--RQ2), while also showing that the association between information load and engagement is not invariant across descriptive norm regimes (RQ4). We interpret these norm-contingent patterns as describing how the informational environment conditions cue use at the threshold stage, rather than as a direct test of normative influence on participation.

\subsection{Engagement Allocation (Like vs. Repost vs. Quote)}

\begin{figure}[htbp]
    \centering
    \includegraphics[width=\linewidth]{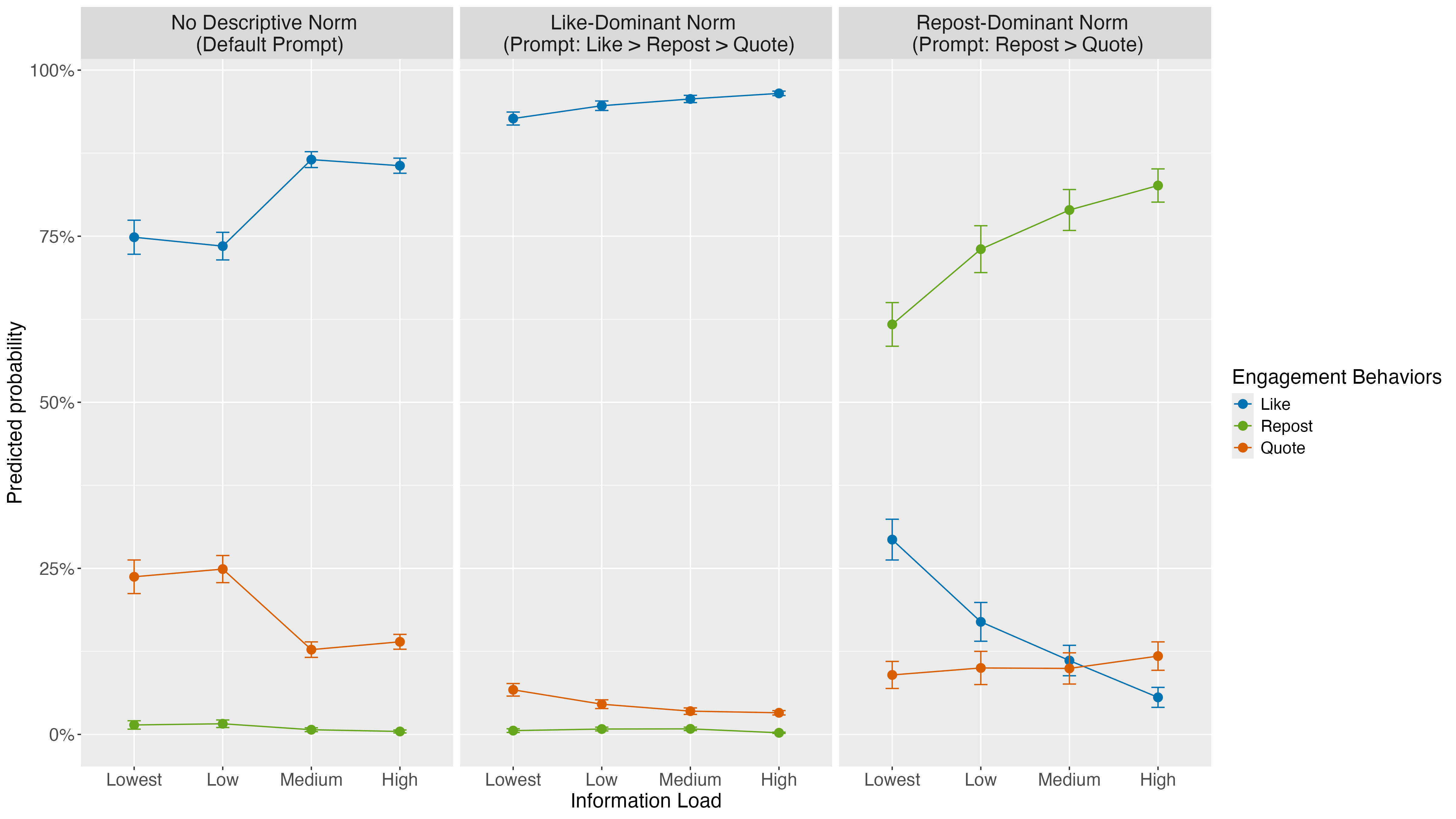}
    \caption{Predicted probabilities of liking, reposting, and quoting across normative regimes and information load conditions.}
    \label{fig:like_repost_quote_prob1}
    
    \vspace{0.5em}
    \small
    \textit{Note.} Estimates are based on a multinomial logistic regression model including all two- and three-way interactions. Error bars represent 95\% confidence intervals.
\end{figure}

\begin{figure}[htbp]
    \centering
    \includegraphics[width=\linewidth]{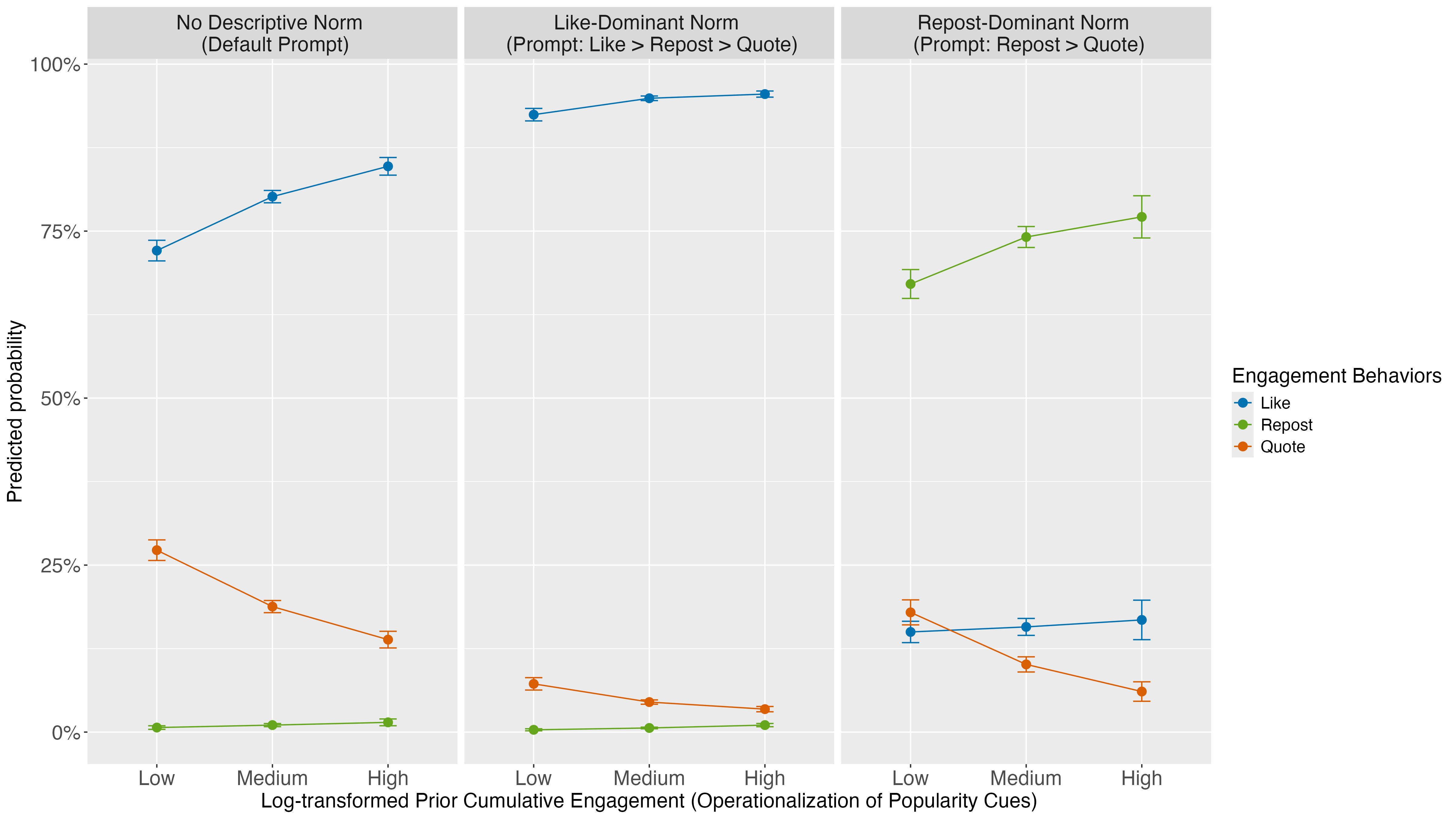}
    \caption{Predicted probabilities of liking, reposting, and quoting across normative regimes and popularity cues.}
    \label{fig:like_repost_quote_prob2}
    
    \vspace{0.5em}
    \small
    \textit{Note.} Estimates are based on a multinomial logistic regression model including all two- and three-way interactions. Error bars represent 95\% confidence intervals.
\end{figure}

The analyses above focus on the engagement threshold between reading and any engagement. We next examine engagement allocation---conditional on engagement, which form agents selected (like, repost, or quote)---using multinomial logistic regression. Model estimates are summarized in \autoref{tab:multinom}. To aid interpretation, \autoref{fig:like_repost_quote_prob1} plots predicted probabilities across descriptive norm regimes and information-load conditions, and \autoref{fig:like_repost_quote_prob2} plots predicted probabilities across descriptive norm regimes and levels of popularity cues.

\begin{table}[htbp]
\centering
\begin{threeparttable}
\caption{Multinomial logistic regression of engagement allocation on popularity cues, information load, and descriptive norm regime.}
\label{tab:multinom}
\scriptsize
\setlength{\tabcolsep}{3pt}
\begin{tabular}{p{5.8cm}rrrrrrrr}
\toprule
& \multicolumn{4}{c}{\textbf{Quote (vs.\ Like)}} & \multicolumn{4}{c}{\textbf{Repost (vs.\ Like)}} \\
\cmidrule(lr){2-5}\cmidrule(lr){6-9}
\textbf{Predictor} & $B$ & $SE$ & OR & $p$ & $B$ & $SE$ & OR & $p$ \\
\midrule
Intercept & -0.180 & 0.081 & 0.835 & .027 & -4.108 & 0.380 & 0.016 & $< .001$ \\
Popularity cues \textsuperscript{a} & -0.935 & 0.098 & 0.392 & $< .001$ & 0.141 & 0.338 & 1.152 & .676 \\
Low Load (vs.\ Lowest Load) \textsuperscript{b} & -0.418 & 0.112 & 0.658 & $< .001$ & -0.084 & 0.498 & 0.919 & .865 \\
Medium Load (vs.\ Lowest Load) \textsuperscript{b} & -1.522 & 0.122 & 0.218 & $< .001$ & 0.262 & 0.537 & 1.300 & .626 \\
High Load (vs.\ Lowest Load) \textsuperscript{b} & -1.683 & 0.116 & 0.186 & $< .001$ & -0.595 & 0.528 & 0.552 & .259 \\
Like-dominant Norm (vs.\ No Norm) \textsuperscript{c} & -1.687 & 0.141 & 0.185 & $< .001$ & -1.773 & 0.697 & 0.170 & .011 \\
Repost-dominant Norm (vs.\ No Norm) \textsuperscript{c} & -0.542 & 0.177 & 0.582 & .002 & 4.490 & 0.395 & 89.100 & $< .001$ \\

\midrule
\multicolumn{9}{l}{\textit{Two-way interactions}} \\
\midrule
Popularity cues $\times$ Low Load & 0.467 & 0.124 & 1.596 & $< .001$ & 0.515 & 0.566 & 1.674 & .363 \\
Popularity cues $\times$ Medium Load & 0.731 & 0.128 & 2.076 & $< .001$ & -1.714 & 0.576 & 0.180 & .003 \\
Popularity cues $\times$ High Load & 0.982 & 0.120 & 2.671 & $< .001$ & -2.765 & 0.585 & 0.063 & $< .001$ \\
Popularity cues $\times$ Like-dominant Norm & 0.204 & 0.154 & 1.226 & .185 & -0.135 & 0.821 & 0.874 & .870 \\
Popularity cues $\times$ Repost-dominant Norm & 0.477 & 0.192 & 1.611 & .013 & -1.827 & 0.483 & 0.161 & $< .001$ \\
Low Load $\times$ Like-dominant Norm & -0.308 & 0.174 & 0.735 & .076 & 1.232 & 0.787 & 3.428 & .118 \\
Medium Load $\times$ Like-dominant Norm & -0.237 & 0.161 & 0.789 & .140 & 0.837 & 0.772 & 2.309 & .278 \\
High Load $\times$ Like-dominant Norm & 0.217 & 0.144 & 1.242 & .131 & 0.521 & 0.751 & 1.684 & .489 \\
Low Load $\times$ Repost-dominant Norm & 0.570 & 0.214 & 1.768 & .008 & 0.828 & 0.532 & 2.290 & .120 \\
Medium Load $\times$ Repost-dominant Norm & 1.179 & 0.206 & 3.252 & $< .001$ & 0.598 & 0.530 & 1.818 & .259 \\
High Load $\times$ Repost-dominant Norm & 1.368 & 0.191 & 3.929 & $< .001$ & 1.451 & 0.522 & 4.267 & .006 \\

\midrule
\multicolumn{9}{l}{\textit{Three-way interactions}} \\
\midrule
Popularity cues $\times$ Low Load $\times$ Like-dominant Norm & 0.512 & 0.218 & 1.668 & .019 & -0.677 & 0.990 & 0.508 & .493 \\
Popularity cues $\times$ Medium Load $\times$ Like-dominant Norm & 0.582 & 0.188 & 1.790 & .002 & 1.358 & 0.975 & 3.890 & .165 \\
Popularity cues $\times$ High Load $\times$ Like-dominant Norm & 0.191 & 0.184 & 1.211 & .300 & 2.007 & 0.952 & 7.440 & .035 \\
Popularity cues $\times$ Low Load $\times$ Repost-dominant Norm & -0.858 & 0.299 & 0.424 & .004 & 0.697 & 0.571 & 2.008 & .222 \\
Popularity cues $\times$ Medium Load $\times$ Repost-dominant Norm & -1.246 & 0.270 & 0.288 & $< .001$ & 3.167 & 0.580 & 23.728 & $< .001$ \\
Popularity cues $\times$ High Load $\times$ Repost-dominant Norm & -1.515 & 0.246 & 0.220 & $< .001$ & 4.334 & 0.574 & 76.282 & $< .001$ \\
\bottomrule
\end{tabular}

\begin{tablenotes}
\vspace{0.5em}
\footnotesize
\item Note. The reference outcome is \textit{like}. OR = odds ratio ($e^B$). $p$ values are two-tailed.
\item[\textsuperscript{a}] Popularity cues is operationalized as the log-transformed cumulative engagement a post had accumulated prior to the focal decision.
\item[\textsuperscript{b}] Information load reflects the number of posts shown to agents; the reference category is the lowest-load condition.
\item[\textsuperscript{c}] Descriptive norm regimes specify prevalence information about engagement forms; the reference category is the no-norm condition.
\end{tablenotes}
\end{threeparttable}
\end{table}

Descriptive norms produced distinct allocation hierarchies. In the no-norm condition, liking was the dominant engagement form, quoting was the secondary form, and reposting remained rare. Under the lowest information load, the predicted probability of liking was 74.8\% (vs.\ 23.7\% for quoting), and under medium information load liking increased to 86.5\% as quoting declined to 12.8\%, with reposting remaining below 2\% throughout. In the like-dominant regime, allocation concentrated strongly on liking across all information-load conditions (92.7\% under the lowest load and 96.5\% under high load), with quoting and reposting both remaining below 7\% and 1\%, respectively. In the repost-dominant regime, reposting became the dominant engagement form across information-load conditions, increasing from 61.7\% under the lowest load to 82.6\% under high load, while liking declined from 29.3\% to 5.6\%.

Information load therefore primarily affected the degree of concentration within regimes: higher load increased reliance on the focal action (liking in the no-norm and like-dominant regimes; reposting in the repost-dominant regime) and reduced the probability of competing forms.

Popularity cues showed a similar concentrating tendency. In the no-norm regime, liking increased from 72.1\% at low popularity cues to 84.7\% at high popularity cues, while quoting declined from 27.2\% to 13.8\% (reposting remained below 2\%). In the like-dominant regime, popularity cues produced only modest additional concentration on liking (92.4\% at low cues vs.\ 95.5\% at high cues), consistent with ceiling effects. In the repost-dominant regime, stronger popularity cues increased reposting from 67.1\% to 77.1\% and reduced quoting from 17.9\% to 6.1\%, suggesting that popularity cues are associated with greater concentration on the focal engagement form within that descriptive norm regime.

These allocation results directly address RQ3 by showing that descriptive norms systematically reorganize engagement allocation toward the normatively emphasized action, while not mechanically enforcing the stated proportions. They also inform RQ4 by showing that information load and popularity cues tend to intensify this reallocation by concentrating engagement on the focal action within each descriptive norm regime. Finally, the popularity-cue patterns complement RQ2 at the allocation stage by showing that popularity cues are associated not only with engagement likelihood (as shown in the threshold model) but also with how engagement is distributed across forms once agents choose to engage.

\section{Conclusions and Discussion}
This study used an LLM-driven social media simulation to examine whether well-established engagement mechanisms can be observed when \textit{information load} and \textit{descriptive norms} are varied experimentally and \textit{popularity cues} are allowed to develop endogenously. The broader aim was not to replace behavioral trace data \citep{salganikBitBitSocial2019, vanatteveldtWhenCommunicationMeets2018}, but to evaluate what an LLM-based simulation can contribute when the contextual features that are typically unobserved or endogenous on platforms are made observable and manipulable by design \citep{waldherrWorldsAgentsProspects2021}. Across models that distinguish engagement thresholds from engagement allocation, we find that engagement responds systematically to information load, descriptive norms, and popularity cues, and that these influences are not constant across environments.

\autoref{tab:finding_summary} summarizes the key findings for each research question and indicates where each mechanism operates within the two-stage engagement process. Rather than restating each result, we highlight two cross-cutting patterns. First, the mechanisms map onto the two stages in distinct ways: information load is most directly reflected in the engage-versus-read threshold, descriptive norms are most evident in how engagement is allocated across response forms, and popularity cues are associated with both stages---most clearly with engagement likelihood at the threshold stage, and with greater concentration on focal actions conditional on engagement. Second, the patterns are conditional rather than additive: the association between popularity cues and engagement varies across levels of information load and across descriptive norm regimes, and higher load tends to concentrate engagement on focal actions within each regime. Together, these features provide a concise basis for interpreting the results and situating their theoretical and methodological implications.

\begin{table}[htbp]
\centering
\caption{Summary of Core Findings Across Research Questions}
\label{tab:finding_summary}
\footnotesize
\begin{tabular}{p{1cm} p{2.5cm} p{2.8cm} p{6.2cm} p{3.0cm}}
\toprule
\textbf{RQ} & \textbf{Mechanism focus} & \textbf{Engagement stage} & \textbf{Core finding} & \textbf{Implication} \\
\midrule
RQ1 & Information load &
Threshold (engage vs.\ read) &
Higher information load generally reduces the likelihood of transitioning from reading to any engagement, especially in the no-norm and like-dominant regimes. &
Participation thresholds are load-sensitive. \\
\addlinespace

RQ2 & Popularity cues / bandwagon effects &
Threshold (primary) and allocation (secondary) &
Stronger popularity cues substantially increase the likelihood of engagement, consistent with bandwagon effects; conditional on engagement, popularity cues are associated with greater concentration on focal actions within regimes. &
Popularity cues shape both participation and allocation. \\
\addlinespace

RQ3 & Descriptive norms &
Allocation (like/repost/quote) &
Descriptive norm prompts systematically reorganize engagement allocation toward the normatively emphasized action without enforcing fixed proportions. &
Norms reorganize engagement composition. \\
\addlinespace

RQ4 & Joint effects &
Threshold and allocation &
Information load, descriptive norms, and popularity cues interact: the association between popularity cues and engagement varies across norm regimes and load levels, with the repost-dominant regime showing crossover patterns in which higher load can amplify responsiveness to popularity cues. &
Mechanisms are conditional, not additive. \\
\bottomrule
\end{tabular}
\end{table}

\subsection{Consistency and Refinements: What Our Findings Imply Theoretically}

Several patterns align with widely tested theories of attention, social influence, and normative behavior. Engagement is context-dependent in visibility-mediated environments: information load, descriptive norms, and popularity cues jointly define the decision setting. They shape how many alternatives compete for attention, which engagement forms appear typical, and which posts appear validated by others. This matters because likes, reposts, and quotes are not interchangeable on most platforms. When the distribution of engagement forms changes, the visibility signals that accumulate also change, with downstream consequences for what becomes more visible and more likely to diffuse. From this perspective, engagement composition is not only a descriptive detail; it is part of how attention constraints, norms, and popularity cues translate into distinct forms of response.

The results also provide a concrete way to think about information load. Rather than treating overload as a general background condition, the findings point to the size of the choice set at the moment of decision as a key theoretical object. Information load is associated with lower engagement likelihood in the baseline and like-dominant regimes, and it also conditions how strongly popularity cues are associated with the engage-versus-read decision. This pattern is consistent with capacity-based accounts \citep{miller_magical_1956, lang_limited_2000} and with social influence theories \citep{bikhchandani_theory_1992, muchnik_social_2013, david_easley_networks_2010} that treat popularity cues as decision-relevant signals whose influence can depend on attentional conditions.

At the same time, several patterns are less emphasized in existing engagement research and suggest refinements. First, the association between popularity cues and engagement is not constant across environments: responsiveness to popularity cues varies with information load and descriptive norm regime \citep{david_easley_networks_2010, muchnik_social_2013}. In particular, the repost-dominant regime produces crossover patterns in which higher information load can suppress engagement when popularity cues are weak but amplify responsiveness once popularity cues reach moderate levels. Second, descriptive norms appear to operate beyond engagement allocation. Even when framed as relative frequencies among like, repost, and quote, prevalence information can shift the engage-versus-read threshold, indicating a spillover from allocation to participation. Third, the no-norm condition is not a neutral reference point. It exhibits a distinctive baseline hierarchy in which liking dominates, quoting remains nontrivial, and reposting is rare. Notably, this baseline ordering aligns with the direction emphasized in the like-dominant regime, suggesting that the like-dominant prompt may primarily amplify an existing default rather than introduce a new ordering. Substantively, these patterns imply that baseline engagement composition may reflect implicit affordance meanings and effort differences that descriptive norms can strengthen, weaken, or redirect depending on the direction of the norm cue.

Because these conclusions are based on agent simulation rather than direct observation of human behavior, they should be treated as theory-relevant patterns that motivate targeted empirical tests rather than as population claims. Their theoretical value lies in clarifying which engagement regularities appear robust under controlled manipulation and which depend on features of the informational environment that are often unmeasured or endogenous in observational research.

\subsection{Validation and Leverage: What Our Study Contributes Methodologically}

Methodologically, this study advances a mechanism-oriented approach to evaluating LLM-driven social simulation. The goal is not to claim behavioral equivalence between artificial agents and human users, but to assess whether an LLM-driven environment can be used to probe well-established mechanisms under controlled conditions. As LLM outputs become increasingly fluent, resemblance in tone or content is not a sufficient basis for inference. This concern is consistent with recent evaluation work showing that even for seemingly basic domains such as commonsense reasoning, “ground truth” can mask substantial heterogeneity in human judgments and that model evaluation is more informative when tied to empirical correspondence with human populations rather than surface plausibility alone \citep{nguyenLargescaleEvaluationCommonsense2026}. A more consequential question is whether simulated behavior responds to systematic changes in the environment in ways that are theoretically interpretable.

A first contribution is to make validation criteria more explicit. Many evaluations of LLM simulation rely on distributional plausibility---for example, whether aggregate engagement rates look reasonable or whether generated content resembles platform discourse. In contrast, the approach used here treats validation as a form of construct testing. A simulation is informative to the extent that it reproduces directional patterns and conditionality implied by established mechanisms when contextual features are manipulated \citep{waldherrWorldsAgentsProspects2021, kozlowskiSimulatingSubjectsPromise2025}. This standard is more demanding than surface realism because it requires consistent mappings between contextual cues and behavioral outcomes, including whether effects change across environments. The implication is that evaluation becomes an empirical test of mechanism sensitivity rather than a judgment about how human-like the output appears \citep{li_simulating_2025}.

A second contribution is to treat prompting as a manipulable informational context with testable implications. Here, descriptive norm prompts function as prevalence information about typical engagement forms, analogous to descriptive norm messages used in behavioral interventions \citep{allcottSocialNormsEnergy2011, papakonstantinouSystematicReviewMetaanalysis2025}. This framing shifts prompts from an implementation detail to an explicit contextual manipulation whose consequences can be evaluated empirically. It also provides a simple diagnostic: if agents are merely following an instruction, norm effects should be relatively stable across conditions; if agents use prompts as contextual prevalence information, norm effects should vary with information load and popularity cues. In our results, descriptive norm effects vary across information-load conditions and levels of popularity cues, and the threshold model shows that the association between popularity cues and engagement changes across descriptive norm regimes. A related design implication is that researchers should be cautious about treating a default prompt as a neutral baseline. Default prompts are not blank prompts: even without explicit descriptive norm information, they contain task framing and implicit expectations that can shape baseline engagement composition. An explicit no-norm baseline therefore helps distinguish prompts that introduce new allocation patterns from prompts that amplify existing defaults.

A third contribution concerns the integration of experimental control with endogenous popularity cues. In observational settings, information load, popularity, and algorithmic visibility are often intertwined and partially unobserved, complicating identification. Many experiments address this problem by manipulating popularity cues exogenously, but doing so can detach the cue from the cumulative processes that generate it in visibility-mediated systems. The present design retains the feedback loop by allowing cumulative engagement counts---operationalized as popularity cues---to emerge from interaction history within each run, while holding network structure, activation rates, and recommendation logic constant across conditions. This makes it possible to examine bandwagon-relevant responsiveness in a setting where popularity cues are produced by the system itself rather than imposed as a static treatment.

A final contribution is to emphasize multi-condition testing and the theory-generative role of LLM simulation. Evaluation under a single parameterization can be misleading because apparent validation may reflect a narrow alignment between one prompt regime and one informational environment. By varying information load and descriptive norms while holding other components constant, the design enables structured stress testing that distinguishes generalizable mechanism signatures from regime-specific artifacts \citep{bail_can_2024}. In this study, for example, the repost-dominant regime revealed crossover patterns in the threshold stage that would not be visible from a single-condition evaluation.

This multi-condition logic also suggests two broader forms of methodological leverage. First, it offers a path toward more controlled cross-platform comparison. Platforms differ in affordances, implicit rules, and descriptive norms about what counts as appropriate engagement, but observational comparisons are difficult because trace data are rarely comparable across platforms and many contextual features shift simultaneously \citep{matassi_agenda_2021}. LLM-driven simulation provides a complementary approach by treating platform-specific norms and affordance meanings as manipulable features of the decision setting \citep{tornbergSimulatingSocialMedia2023}. By holding constant core elements of the environment (e.g., content pool, network structure, activity rate, and recommendation logic) while varying descriptive norm regimes that approximate platform-specific engagement climates, researchers can conduct controlled comparisons of how platform contexts reorganize engagement thresholds and engagement allocation.

Second, LLM-based simulation can complement online field experiments. Online field experiments provide strong causal leverage by intervening in naturalistic settings, but they also face constraints that are difficult to avoid in platform contexts, including ethical and legal concerns, construct validity challenges, and interference or spillover across connected units \citep{muiseOnlineFieldExperiments2019}. Simulation addresses a different part of the research pipeline: it enables controlled variation in information load, descriptive norms, and popularity cues without intervening on real users, and it can be used to probe whether hypothesized mechanisms are robust across environments before investing in costly or sensitive deployments. In these ways, simulation is best viewed as a complement to human data: it can sharpen hypotheses, clarify manipulable levers, and surface boundary conditions that can later be adjudicated with field experiments, lab studies, or observational designs.

Taken together, the study illustrates how LLM-driven simulation can support controlled counterfactual reasoning about engagement mechanisms in settings where key contextual features are otherwise difficult to observe and endogenous to platform dynamics.

\subsection{Limitations and Future Research}

Several limitations qualify the scope of the present inferences. First, the simulation focuses on engagement behaviors that directly affect feed-level visibility (like, repost, quote) and does not model commenting. Comments may respond to different incentives than redistribution-oriented actions and can shape conversational norms even when they contribute less to feed-level propagation. Incorporating comment behavior would allow future work to compare mechanisms shaping visibility-producing engagement versus conversational participation.

Second, the simulation is anchored in a single topical domain (DeepSeek-related discussions). Topic specificity can shape baseline engagement hierarchies, the meaning of quoting versus reposting, and the interpretability of descriptive norm cues. Future research should extend the design to multi-issue environments in which agents face competing issue streams within the same feed, allowing researchers to examine how users allocate attention across issues and decide not only \emph{how} to engage, but \emph{which} issue-specific messages warrant engagement under varying information load, descriptive norm regimes, and popularity cues.

Third, agent activity is controlled by fixing the probability that users are active at each timestep. This improves comparability across conditions but abstracts away from endogenous attention cycles and bursty participation that characterize real platforms. Allowing activity to vary endogenously---for example, as a function of popularity cues, information load, or time-of-day rhythms---would provide a stronger test of whether the same mechanisms shape participation timing and intensity.

Fourth, the results may depend on the specific LLM configuration and decoding settings used here (model family, temperature, and prompt structure). Testing whether the same mechanism signatures replicate across model families and parameterizations is necessary to distinguish general properties of LLM-driven simulation from model-specific artifacts.

Relatedly, some allocation patterns may reflect model-specific response preferences rather than platform norms alone. Because quoting requires generating additional text whereas reposting is comparatively minimal, an underlying tendency to favor longer outputs could tilt engagement allocation toward more elaborated responses in some settings. Prior work documents a ``verbosity bias'' in LLM-based preference labeling, where longer responses can be favored even when quality is similar \citep{saito_verbosity_2023}. Future work can test this possibility directly by imposing length constraints, using templated short quotes, or comparing models with different alignment and reward-training histories.

Finally, content visibility is generated by a fixed recommendation logic that is held constant across conditions. While this supports identification, alternative ranking policies could change the salience of popularity cues and the distribution of posts agents encounter. Future work can treat recommendation rules as additional experimental levers to evaluate how algorithmic selection and ranking condition load sensitivity, bandwagon-relevant responsiveness, and engagement allocation.

\printbibliography

@misc{yang_oasis_2024,
    title = {{OASIS}: {Open} {Agent} {Social} {Interaction} {Simulations} with {One} {Million} {Agents}},
    shorttitle = {{OASIS}},
    url = {http://arxiv.org/abs/2411.11581},
    doi = {10.48550/arXiv.2411.11581},
    urldate = {2025-01-09},
    publisher = {arXiv},
    author = {Yang, Ziyi and Zhang, Zaibin and Zheng, Zirui and Jiang, Yuxian and Gan, Ziyue and Wang, Zhiyu and Ling, Zijian and Chen, Jinsong and Ma, Martz and Dong, Bowen and Gupta, Prateek and Hu, Shuyue and Yin, Zhenfei and Li, Guohao and Jia, Xu and Wang, Lijun and Ghanem, Bernard and Lu, Huchuan and Lu, Chaochao and Ouyang, Wanli and Qiao, Yu and Torr, Philip and Shao, Jing},
    month = nov,
    year = {2024},
    note = {},
}

@misc{li_simulating_2025,
    title = {Simulating {Society} {Requires} {Simulating} {Thought}},
    url = {http://arxiv.org/abs/2506.06958},
    doi = {10.48550/arXiv.2506.06958},
    urldate = {2025-12-13},
    publisher = {arXiv},
    author = {Li, Chance Jiajie and Wu, Jiayi and Mo, Zhenze and Qu, Ao and Tang, Yuhan and Zhao, Kaiya Ivy and Gan, Yulu and Fan, Jie and Yu, Jiangbo and Zhao, Jinhua and Liang, Paul and Alonso, Luis and Larson, Kent},
    month = oct,
    year = {2025},
    note = {},
}

@book{david_easley_networks_2010,
    address = {New York},
    title = {Networks, {Crowds}, and {Markets}: {Reasoning} {About} a {Highly} {Connected} {World}},
    publisher = {Cambridge University Press},
    author = {David Easley and Jon Kleinberg},
    year = {2010},
}

@article{demszky_using_2023,
    title = {Using large language models in psychology},
    copyright = {2023 Springer Nature America, Inc.},
    issn = {2731-0574},
    url = {https://www.nature.com/articles/s44159-023-00241-5},
    doi = {10.1038/s44159-023-00241-5},
    language = {en},
    urldate = {2023-10-14},
    journal = {Nature Reviews Psychology},
    publisher = {Nature Publishing Group},
    author = {Demszky, Dorottya and Yang, Diyi and Yeager, David S. and Bryan, Christopher J. and Clapper, Margarett and Chandhok, Susannah and Eichstaedt, Johannes C. and Hecht, Cameron and Jamieson, Jeremy and Johnson, Meghann and Jones, Michaela and Krettek-Cobb, Danielle and Lai, Leslie and JonesMitchell, Nirel and Ong, Desmond C. and Dweck, Carol S. and Gross, James J. and Pennebaker, James W.},
    month = oct,
    year = {2023},
    pages = {1--14},
}

@article{cui_large-scale_2025,
    title = {A large-scale replication of scenario-based experiments in psychology and management using large language models},
    copyright = {2025 The Author(s), under exclusive licence to Springer Nature America, Inc.},
    issn = {2662-8457},
    url = {https://www.nature.com/articles/s43588-025-00840-7},
    doi = {10.1038/s43588-025-00840-7},
    language = {en},
    urldate = {2025-08-20},
    journal = {Nature Computational Science},
    publisher = {Nature Publishing Group},
    author = {Cui, Ziyan and Li, Ning and Zhou, Huaikang},
    month = jul,
    year = {2025},
    pages = {1--8},
}

@article{bail_can_2024,
    title = {Can {Generative} {AI} improve social science?},
    volume = {121},
    url = {https://www-pnas-org.proxy2.cl.msu.edu/doi/10.1073/pnas.2314021121},
    doi = {10.1073/pnas.2314021121},
    number = {21},
    urldate = {2024-06-07},
    journal = {Proceedings of the National Academy of Sciences},
    publisher = {Proceedings of the National Academy of Sciences},
    author = {Bail, Christopher A.},
    month = may,
    year = {2024},
    pages = {e2314021121},
}

@article{rodriguez_quantifying_2014,
    title = {Quantifying {Information} {Overload} in {Social} {Media} and {Its} {Impact} on {Social} {Contagions}},
    volume = {8},
    copyright = {Copyright (c) 2021 Proceedings of the International AAAI Conference on Web and Social Media},
    issn = {2334-0770},
    url = {https://ojs.aaai.org/index.php/ICWSM/article/view/14549},
    doi = {10.1609/icwsm.v8i1.14549},
    language = {en},
    number = {1},
    urldate = {2026-02-07},
    journal = {Proceedings of the International AAAI Conference on Web and Social Media},
    author = {Rodriguez, Manuel Gomez and Gummadi, Krishna and Schoelkopf, Bernhard},
    month = may,
    year = {2014},
    note = {},
    pages = {170--179},
}

@article{weng_competition_2012,
    title = {Competition among memes in a world with limited attention},
    volume = {2},
    issn = {2045-2322},
    url = {https://www.nature.com/articles/srep00335},
    doi = {10.1038/srep00335},
    language = {en},
    number = {1},
    urldate = {2026-02-07},
    journal = {Scientific Reports},
    author = {Weng, L. and Flammini, A. and Vespignani, A. and Menczer, F.},
    month = mar,
    year = {2012},
    note = {},
    pages = {335},
}

@article{feng_competing_2015,
    title = {Competing for {Attention} in {Social} {Media} under {Information} {Overload} {Conditions}},
    volume = {10},
    issn = {1932-6203},
    url = {https://journals.plos.org/plosone/article?id=10.1371/journal.pone.0126090},
    doi = {10.1371/journal.pone.0126090},
    language = {en},
    number = {7},
    urldate = {2026-02-07},
    journal = {PLOS ONE},
    publisher = {Public Library of Science},
    author = {Feng, Ling and Hu, Yanqing and Li, Baowen and Stanley, H. Eugene and Havlin, Shlomo and Braunstein, Lidia A.},
    month = jul,
    year = {2015},
    note = {},
    pages = {e0126090},
}

@article{miller_magical_1956,
    address = {Washington, US},
    title = {The magical number seven, plus or minus two: {Some} limits on our capacity for processing information},
    volume = {63},
    copyright = {© 1956, American Psychological Association. All rights, including for text and data mining, AI training, and similar technologies, are reserved.},
    issn = {0033-295X},
    shorttitle = {The magical number seven, plus or minus two},
    url = {https://www.proquest.com/docview/614291982/abstract/5BF95EB0F3984668PQ/1},
    doi = {10.1037/h0043158},
    language = {English},
    number = {2},
    urldate = {2026-02-07},
    journal = {Psychological Review},
    publisher = {American Psychological Association},
    author = {Miller, George A.},
    month = mar,
    year = {1956},
    note = {},
    pages = {81--97},
}

@article{lang_limited_2000,
    title = {The {Limited} {Capacity} {Model} of {Mediated} {Message} {Processing}},
    volume = {50},
    issn = {0021-9916},
    url = {https://doi.org/10.1111/j.1460-2466.2000.tb02833.x},
    doi = {10.1111/j.1460-2466.2000.tb02833.x},
    number = {1},
    urldate = {2026-02-07},
    journal = {Journal of Communication},
    author = {Lang, Annie},
    month = mar,
    year = {2000},
    note = {},
    pages = {46--70},
}

@article{bikhchandani_theory_1992,
    title = {A {Theory} of {Fads}, {Fashion}, {Custom}, and {Cultural} {Change} as {Informational} {Cascades}},
    volume = {100},
    issn = {0022-3808},
    url = {https://www.journals.uchicago.edu/doi/abs/10.1086/261849},
    doi = {10.1086/261849},
    number = {5},
    urldate = {2026-02-08},
    journal = {Journal of Political Economy},
    publisher = {The University of Chicago Press},
    author = {Bikhchandani, Sushil and Hirshleifer, David and Welch, Ivo},
    month = oct,
    year = {1992},
    note = {},
    pages = {992--1026},
}

@article{muchnik_social_2013,
    title = {Social {Influence} {Bias}: {A} {Randomized} {Experiment}},
    volume = {341},
    shorttitle = {Social {Influence} {Bias}},
    url = {https://www.science.org/doi/10.1126/science.1240466},
    doi = {10.1126/science.1240466},
    number = {6146},
    urldate = {2026-02-08},
    journal = {Science},
    publisher = {American Association for the Advancement of Science},
    author = {Muchnik, Lev and Aral, Sinan and Taylor, Sean J.},
    month = aug,
    year = {2013},
    note = {},
    pages = {647--651},
}

@article{cialdini_focus_1990,
    address = {US},
    title = {A focus theory of normative conduct: {Recycling} the concept of norms to reduce littering in public places},
    volume = {58},
    issn = {1939-1315},
    shorttitle = {A focus theory of normative conduct},
    doi = {10.1037/0022-3514.58.6.1015},
    number = {6},
    journal = {Journal of Personality and Social Psychology},
    publisher = {American Psychological Association},
    author = {Cialdini, Robert B. and Reno, Raymond R. and Kallgren, Carl A.},
    year = {1990},
    note = {},
    pages = {1015--1026},
}

@misc{yangQwen3TechnicalReport2025,
    title = {Qwen3 {Technical} {Report}},
    url = {http://arxiv.org/abs/2505.09388},
    doi = {10.48550/arXiv.2505.09388},
    urldate = {2026-02-21},
    publisher = {arXiv},
    author = {Yang, An and Li, Anfeng and Yang, Baosong and Zhang, Beichen and Hui, Binyuan and Zheng, Bo and Yu, Bowen and Gao, Chang and Huang, Chengen and Lv, Chenxu and Zheng, Chujie and Liu, Dayiheng and Zhou, Fan and Huang, Fei and Hu, Feng and Ge, Hao and Wei, Haoran and Lin, Huan and Tang, Jialong and Yang, Jian and Tu, Jianhong and Zhang, Jianwei and Yang, Jianxin and Yang, Jiaxi and Zhou, Jing and Zhou, Jingren and Lin, Junyang and Dang, Kai and Bao, Keqin and Yang, Kexin and Yu, Le and Deng, Lianghao and Li, Mei and Xue, Mingfeng and Li, Mingze and Zhang, Pei and Wang, Peng and Zhu, Qin and Men, Rui and Gao, Ruize and Liu, Shixuan and Luo, Shuang and Li, Tianhao and Tang, Tianyi and Yin, Wenbiao and Ren, Xingzhang and Wang, Xinyu and Zhang, Xinyu and Ren, Xuancheng and Fan, Yang and Su, Yang and Zhang, Yichang and Zhang, Yinger and Wan, Yu and Liu, Yuqiong and Wang, Zekun and Cui, Zeyu and Zhang, Zhenru and Zhou, Zhipeng and Qiu, Zihan},
    month = may,
    year = {2025},
    note = {},
}

@book{salganikBitBitSocial2019,
    title = {Bit by bit: {Social} research in the digital age},
    isbn = {0-691-19610-9},
    publisher = {Princeton University Press},
    author = {Salganik, Matthew J.},
    year = {2019},
}

@article{waldherrWorldsAgentsProspects2021,
    title = {Worlds of {Agents}: {Prospects} of {Agent}-{Based} {Modeling} for {Communication} {Research}},
    volume = {15},
    issn = {1931-2458},
    url = {https://doi.org/10.1080/19312458.2021.1986478},
    doi = {10.1080/19312458.2021.1986478},
    number = {4},
    journal = {Communication Methods and Measures},
    publisher = {Routledge},
    author = {Waldherr, Annie and Hilbert, Martin and González-Bailón, Sandra},
    month = oct,
    year = {2021},
    note = {},
    pages = {243--254},
}

@article{allcottSocialNormsEnergy2011,
    series = {Special {Issue}: {The} {Role} of {Firms} in {Tax} {Systems}},
    title = {Social norms and energy conservation},
    volume = {95},
    issn = {0047-2727},
    url = {https://www.sciencedirect.com/science/article/pii/S0047272711000478},
    doi = {10.1016/j.jpubeco.2011.03.003},
    number = {9},
    urldate = {2026-02-22},
    journal = {Journal of Public Economics},
    author = {Allcott, Hunt},
    month = oct,
    year = {2011},
    note = {},
    pages = {1082--1095},
}

@article{papakonstantinouSystematicReviewMetaanalysis2025,
    title = {A systematic review and meta-analysis of the effectiveness of social norms messaging approaches for improving health behaviours in developed countries},
    volume = {9},
    issn = {2397-3374},
    doi = {10.1038/s41562-025-02275-6},
    language = {eng},
    number = {12},
    journal = {Nature Human Behaviour},
    author = {Papakonstantinou, Trisevgeni and Flecke, Sarah Lynn and Edmunds, C. E. R. and Cross, Rosina and Tran, Anh and Gold, Natalie},
    month = dec,
    year = {2025},
    note = {},
    pages = {2632--2650},
}

@article{kozlowskiSimulatingSubjectsPromise2025,
    title = {Simulating {Subjects}: {The} {Promise} and {Peril} of {Artificial} {Intelligence} {Stand}-{Ins} for {Social} {Agents} and {Interactions}},
    volume = {54},
    issn = {0049-1241},
    shorttitle = {Simulating {Subjects}},
    url = {https://doi.org/10.1177/00491241251337316},
    doi = {10.1177/00491241251337316},
    language = {EN},
    number = {3},
    urldate = {2026-02-22},
    journal = {Sociological Methods \& Research},
    publisher = {SAGE Publications Inc},
    author = {Kozlowski, Austin C. and Evans, James},
    month = aug,
    year = {2025},
    note = {},
    pages = {1017--1073},
}

@article{xieUsingLargeLanguage2025,
    title = {Using large language models to categorize strategic situations and decipher motivations behind human behaviors},
    volume = {122},
    url = {https://www.pnas.org/doi/10.1073/pnas.2512075122},
    doi = {10.1073/pnas.2512075122},
    number = {35},
    urldate = {2025-09-02},
    journal = {Proceedings of the National Academy of Sciences},
    publisher = {Proceedings of the National Academy of Sciences},
    author = {Xie, Yutong and Mei, Qiaozhu and Yuan, Walter and Jackson, Matthew O.},
    month = sep,
    year = {2025},
    pages = {e2512075122},
}

@article{loruSimulationJudgmentLLMs2025,
    title = {The simulation of judgment in {LLMs}},
    volume = {122},
    url = {https://www.pnas.org/doi/10.1073/pnas.2518443122},
    doi = {10.1073/pnas.2518443122},
    number = {42},
    urldate = {2026-01-29},
    journal = {Proceedings of the National Academy of Sciences},
    publisher = {Proceedings of the National Academy of Sciences},
    author = {Loru, Edoardo and Nudo, Jacopo and Di Marco, Niccolò and Santirocchi, Alessandro and Atzeni, Roberto and Cinelli, Matteo and Cestari, Vincenzo and Rossi-Arnaud, Clelia and Quattrociocchi, Walter},
    month = oct,
    year = {2025},
    note = {},
    pages = {e2518443122},
}

@misc{tornbergSimulatingSocialMedia2023,
    title = {Simulating {Social} {Media} {Using} {Large} {Language} {Models} to {Evaluate} {Alternative} {News} {Feed} {Algorithms}},
    url = {http://arxiv.org/abs/2310.05984},
    doi = {10.48550/arXiv.2310.05984},
    urldate = {2023-10-11},
    publisher = {arXiv},
    author = {Törnberg, Petter and Valeeva, Diliara and Uitermark, Justus and Bail, Christopher},
    month = oct,
    year = {2023},
    note = {arXiv:2310.05984 [cs]},
}

@article{vanatteveldtWhenCommunicationMeets2018,
    title = {When {Communication} {Meets} {Computation}: {Opportunities}, {Challenges}, and {Pitfalls} in {Computational} {Communication} {Science}},
    volume = {12},
    issn = {1931-2458;1931-2466},
    doi = {10.1080/19312458.2018.1458084},
    number = {2-3},
    journal = {Communication Methods and Measures},
    author = {van Atteveldt, Wouter and Peng, Tai-Quan},
    year = {2018},
    pages = {81--92},
}

@article{nguyenLargescaleEvaluationCommonsense2026,
    title = {A large-scale evaluation of commonsense knowledge in humans and large language models},
    issn = {2752-6542},
    url = {https://doi.org/10.1093/pnasnexus/pgag029},
    doi = {10.1093/pnasnexus/pgag029},
    urldate = {2026-02-20},
    journal = {PNAS Nexus},
    author = {Nguyen, Tuan Dung and Watts, Duncan J and Whiting, Mark E},
    month = feb,
    year = {2026},
    note = {},
    pages = {pgag029},
}

@article{muiseOnlineFieldExperiments2019,
    title = {Online field experiments},
    volume = {29},
    issn = {0129-2986},
    url = {https://doi.org/10.1080/01292986.2018.1453850},
    doi = {10.1080/01292986.2018.1453850},
    number = {3},
    urldate = {2026-03-04},
    journal = {Asian Journal of Communication},
    publisher = {Routledge},
    author = {Muise, Daniel and Pan, Jennifer},
    month = may,
    year = {2019},
    note = {},
    pages = {217--234},
}

@article{matassi_agenda_2021,
    title = {An {Agenda} for {Comparative} {Social} {Media} {Studies}: {The} {Value} of {Understanding} {Practices} {From} {Cross}-{National}, {Cross}-{Media}, and {Cross}-{Platform} {Perspectives}},
    volume = {15},
    issn = {1932-8036},
    shorttitle = {An {Agenda} for {Comparative} {Social} {Media} {Studies}},
    url = {https://ijoc.org/index.php/ijoc/article/view/15042},
    language = {en},
    urldate = {2026-03-04},
    journal = {International Journal of Communication},
    author = {Matassi, Mora and Boczkowski, Pablo},
    month = jan,
    year = {2021},
    note = {},
    pages = {207--228},
}

@misc{saito_verbosity_2023,
    title = {Verbosity {Bias} in {Preference} {Labeling} by {Large} {Language} {Models}},
    url = {http://arxiv.org/abs/2310.10076},
    doi = {10.48550/arXiv.2310.10076},
    urldate = {2026-03-06},
    publisher = {arXiv},
    author = {Saito, Keita and Wachi, Akifumi and Wataoka, Koki and Akimoto, Youhei},
    month = oct,
    year = {2023},
    note = {},
}

\end{document}